\documentclass{article}
\usepackage{ijcai20}

\usepackage{times}

\usepackage{soul}
\usepackage{url}
\usepackage[hidelinks]{hyperref}
\usepackage[utf8]{inputenc}
\usepackage[small]{caption}
\usepackage{graphicx}
\usepackage{amsmath}
\usepackage{booktabs}

\usepackage{amsthm}
\usepackage{algorithm}
\usepackage{algorithmic}
\usepackage{amssymb}
\usepackage{subcaption}
\usepackage{xcolor}
\usepackage{enumitem}

\newcommand{\NAME}{Ours}
\newcommand{\parareduce}{91.6\%}
\newcommand{\flopreduce}{83.7\%}

\title{Feature Statistics Guided Efficient Filter Pruning}

\author{
Hang Li$^1$\and
Chen Ma$^1$\footnote{Corresponding Author}\and
Wei Xu$^2$\And
Xue Liu$^{1}$\\
\affiliations
$^1$School of Computer Science, McGill University\\ 
$^2$Institute for Interdisciplinary Information Sciences, Tsinghua University\\
\emails
\{hang.li3, chen.ma2\}@mail.mcgill.ca,
weixu@tsinghua.edu.cn,
xueliu@cs.mcgill.ca
}

\begin{document}

\maketitle

\begin{abstract}
Building compact convolutional neural networks (CNNs) with reliable performance is a critical but challenging task, especially when deploying them in real-world applications. As a common approach to reduce the size of CNNs, pruning methods delete part of the CNN filters according to some metrics such as $l1$-norm. However, previous methods hardly leverage the information variance in a single feature map and the similarity characteristics among feature maps. In this paper, we propose a novel filter pruning method, which incorporates two kinds of feature map selections: diversity-aware selection (DFS) and similarity-aware selection (SFS). DFS aims to discover features with low information diversity while SFS removes features that have high similarities with others. We conduct extensive empirical experiments with various CNN architectures on publicly available datasets. The experimental results demonstrate that our model obtains up to \parareduce{} parameter decrease and \flopreduce{} FLOPs reduction with almost no accuracy loss.
\end{abstract}

\section{Introduction}
Deep convolutional neural networks (CNNs) have evolved to the state-of-the-art technique on various tasks, including image classification~\cite{AlexNet}, object detection~\cite{girshick2014rich} and sentence classification~\cite{NLP_CNN}. Due to the parameter sharing and local connectivity schemes, CNNs own the powerful representation and approximation ability that can benefit downstream tasks. For example, the classification accuracy of CNNs in the ImageNet challenge has increased from 84.7\% in 2012 (AlexNet~\cite{AlexNet}) to 96.5\% in 2015 (ResNet-152~\cite{resnet}). 

Although CNNs yield the state-of-the-art performance on various tasks, they still suffer from high storage and computation overheads. Specifically, CNNs cost a huge space to store millions or even billions of parameters; the floating-point operations (FLOPs) of CNNs are intensive since a large quantity of multiplication and addition operations are executed in convolutional layers. These two drawbacks impede deploying CNNs in real-world applications especially when the storage and computation resources are limited.

Filter pruning, as a promising solution to address the aforementioned issues, has drawn significant interests from both academia and industry. The reasons are two-fold. First, most filter pruning methods are conducted on predefined architectures without extra designs~\cite{han2015learning}. Second, filter pruning techniques do not introduce sparsity to architectures like weight pruning methods~\cite{li2016pruning}. In particular, local pruning methods remove less important filters according to the pruning ratios in each layer, which leads to a fixed architecture with finely trained weights. For example,~\cite{li2016pruning} prunes filters with a low $l1$-norm in each layer. However, ~\cite{liu2018rethinking} shows that once the pruned architecture has been defined, the performance relies more on the architecture rather than learned weights, which makes local pruning less effective. Compared with local pruning, global pruning distinguishes the importance of filters across all layers, which can achieve better performance. The pruned architecture is automatically determined by the global pruning algorithm. As a representative approach of global pruning, ~\cite{liu2017learning} imposes a sparsity-induced regularization on the scaling factors of batch normalization layers and prunes those channels with smaller factors. 

Even though previous works have proposed effective methods to compress the CNN architecture, we argue that two factors of feature maps are rarely incorporated. On the one hand, the information variance of a feature map can be a good indicator for discriminating the effectiveness of feature maps. That is, if the values in a feature map do not vary a lot, the amount of information it contains may be limited. On the other hand, the relationships (e.g. similarity) between feature maps play a significant role in preserving effective feature maps. If two features have high similarities, then one of them can be considered as redundant. However, previous works mainly utilize the metrics within a feature map without considering the similarities between feature maps. For example,~\cite{li2016pruning} only applies $l$1-norm to select feature maps. Solely depending on $l$1-norm may keep multiple similar feature maps with the same $l$1-norm value, which will lead to incomplete pruning.

To incorporate the aforementioned intuitions, we propose a framework containing two stpdf of feature map selections to prune less diverse feature maps and corresponding filters. The first step employs the \underline{d}iversity-aware \underline{f}eature \underline{s}election (DFS) to remove feature maps with less information variance. In particular, we apply the mean standard deviation (M-std) of values in a feature map to measure the information variance degree. The feature map with the lowest M-std will be pruned.
The second step is the \underline{s}imilarity-aware \underline{f}eature \underline{s}election (SFS) for deleting feature maps that have high similarities with each others. We compute the cosine similarity among all features and delete the feature whose similarity is larger than a threshold. 
Moreover, we observe that the M-std distribution varies a lot in different layers of ResNet. This motivates us to adopt a fine-grained pruning strategy, which contains two pruning processes working on different parts of a residual block.
We extensively evaluate our method with many state-of-the-art approaches and different metrics on three publicly available datasets. The experimental results show the improvements of our model over other baselines.

Our contributions are summarized as follows:
\begin{itemize}[leftmargin=*]
    \item We introduce an effective filter pruning method to compress CNN models that have a large number of parameters and FLOPs. We employ a diversity-aware feature selection to distinguish informative feature maps and propose a similarity-aware feature selection to remain representative feature maps.
    
    \item Our method is a global filter pruning method that compares the redundancy degree of feature maps across all layers. More importantly, we propose a fine-grained pruning strategy for ResNet, which differs from many existing methods~\cite{he2019filter,li2016pruning,luo2017thinet}.  
    
    \item We extensively evaluate our methods with multiple CNN architectures on three datasets. We show the significantly improved effectiveness of our proposed method, which can reduce parameters of MobileNet by up to \parareduce{} and FLOPs decrease up to \flopreduce{} with limited accuracy drop. 
\end{itemize}{} 

\section{Related Works}
Convolutional neural networks have many parameters to provide enough model capacity, making them both computationally and memory intensive. Pruning methods aim to reduce the number of parameters in a CNN model. Based on the over-parameterization hypothesis~\cite{ba2014deep}, a considerable amount of parameters can be pruned while CNNs maintain a promising accuracy. By reducing the number of weights or filters, we can save the storage space and also reduce the computational complexity and memory footprint of a model during testing.

Weight pruning is a strategy that deletes parameters with small ``saliency".~\cite{lecun1990optimal} introduces the Optimal Brain Damage(OBD) method which is one of the earliest attempts at pruning neural networks. OBD defines the ``saliency" by calculating the second derivative of the objective function concerning the parameters.~\cite{han2015learning} present a generic iterative framework for neural network pruning, in which all weights below a threshold are removed from the network.~\cite{guo2016dynamic} propose a dynamic compression algorithm named Dynamic Network Surgery(DNS) to prune or rebuild the connections during the learning process. Weight pruning is fairly efficient in terms of reducing model size. However, networks compressed by weight pruning methods become sparse, requiring additional sparse libraries or even specialized hardware to run. The filter structure of CNNs allows us to perform filter pruning, which is a naturally structured method without introducing sparsity. And in this paper, we also focus on the filter pruning method.

Filter pruning methods prune the convolutional filters or channels of CNN models which make the deep networks thinner. Based on the assumption that CNNs usually have a significant redundancy among filters, researchers introduce various metrics~\cite{li2016pruning,luo2017thinet,liu2017learning,wang2019cop} to measure the importance of filters, which is the key issue of filter pruning. ~\cite{li2016pruning} measures the relative importance of a filter in each layer by calculating the sum of its absolute kernel weights. Beyond such a magnitude-based method, ~\cite{APoz} proposes that if the majority of the value in a filter is zero, this filter is likely to be redundant. They compute the Average Percentage of Zeros (APoZ) of each filter as its importance score.  

Except for adopting the magnitude of a filter as an important metric, feature maps are also significant. Reconstruction-based methods seek to do filter pruning by minimizing the reconstruction error of feature maps between the pruned model and a pre-trained model.~\cite{luo2017thinet} propose ThiNet which transforms the filter selection problem into the optimization of reconstruction error computed from the next layer's feature map.~\cite{he2017channel}  uses the LASSO regression to obtain a subset of filters that can reconstruct the corresponding output in each layer. Even though this reconstruction-based method considers the feature map information, it has a limitation that the reconstructed feature map might have high similarity which is redundant information.

Besides the aforementioned methods that prune filters after obtaining a trained neural network, there are some explorations~\cite{liu2019metapruning,he2018amc} to get the importance of each filter during the training process. In ~\cite{liu2017learning}, a scaling parameter $\gamma$ is introduced to each channel and is trained simultaneously with the rest of the weights by adding L1-norm of $\gamma$ in the loss function. Channels with small factors are pruned and the network is fine-tuned after pruning. ~\cite{yamamoto2018pcas} inserts a self-attention module in the pre-trained convolutional layer or fully connected layer to learn the importance of each channel.~\cite{luo2018autopruner} learn a 0-1 indicator which can multiply with feature maps as the input of the next layer in a joint training manner.

However, our proposed method is different from previous approaches. We apply a diversity-aware feature selection process to remove features with lower information variance. Besides, a similarity-aware feature selection process is utilized to discover those closely related features.

\section{Methodology}
\begin{figure*}[tb]
\centering
\includegraphics[width=0.9\linewidth]{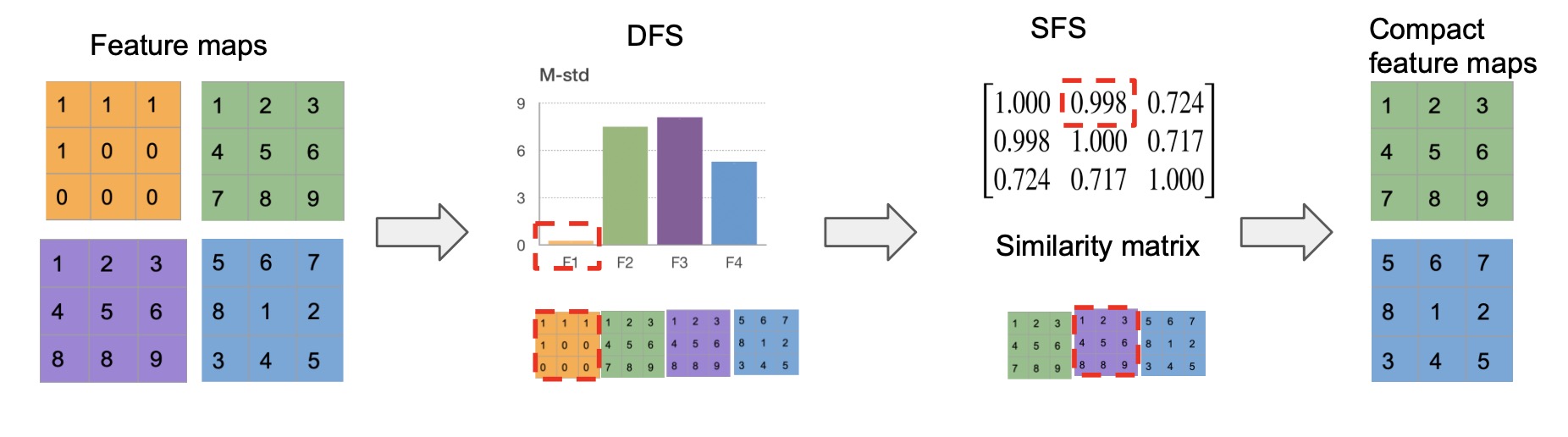}
\caption{The framework of our proposed feature pruning approach. After obtaining feature maps of a trained CNN model, we first do the diversity-aware feature selection (DFS) by removing feature maps with the smaller M-std value. Then the similarity-aware feature selection (SFS) is further used to prune the redundant feature maps with higher cosine similarity.}

\label{fig:pip}
\end{figure*}

\subsection{Preliminaries}
Given a convolutional neural network~(CNN) with $L$ convolution layers, we assume the dimension of the input feature $\mathbf{X}_i$ at the $i_{th}$ layer is $\mathbb{R}^{N_{i} \times W_i \times H_i}$, where $N_i$, $H_i$ and $W_i$ denote the number of channels, rows and columns, respectively. The output dimension at this layer is $\mathbb{R}^{N_{i+1} \times W_{i+1} \times H_{i+1}}$. The corresponding filter set of the $i_{th}$ layer is $\mathcal{F}_i = \{\mathbf{F}_{i,1}, \mathbf{F}_{i,2}, . . . , \mathbf{F}_{i,N_{i+1}}\}$, where $\mathbf{F}_{i,j} \in \mathbb{R}^{N_{i} \times K \times K}$ and $ K \times K$ is the kernel size. The convolutional operation of the $i_{th}$ layer is denoted as:
\begin{equation}
    \mathbf{X}_{i+1,j} = \mathbf{F}_{i,j}\ast \mathbf{X}_{i} \ ,  \quad 1\leqslant j \leqslant N_{i+1},
\end{equation}
where $\mathbf{X}_{i,k} \in \mathbb{R}^{W_i \times H_i}$ represents the feature map of the $k_{th}$ channel at the $i_{th}$ layer.

Given a model \textbf{M} trained on dataset $\{(\hat{x}_i,\hat{y}_i)\}$, our task is to delete redundant feature maps and corresponding filters. 

\subsection{Diversity and Similarity in Feature Maps}

\begin{figure}[tb]
\centering
\includegraphics[width=0.7\linewidth]{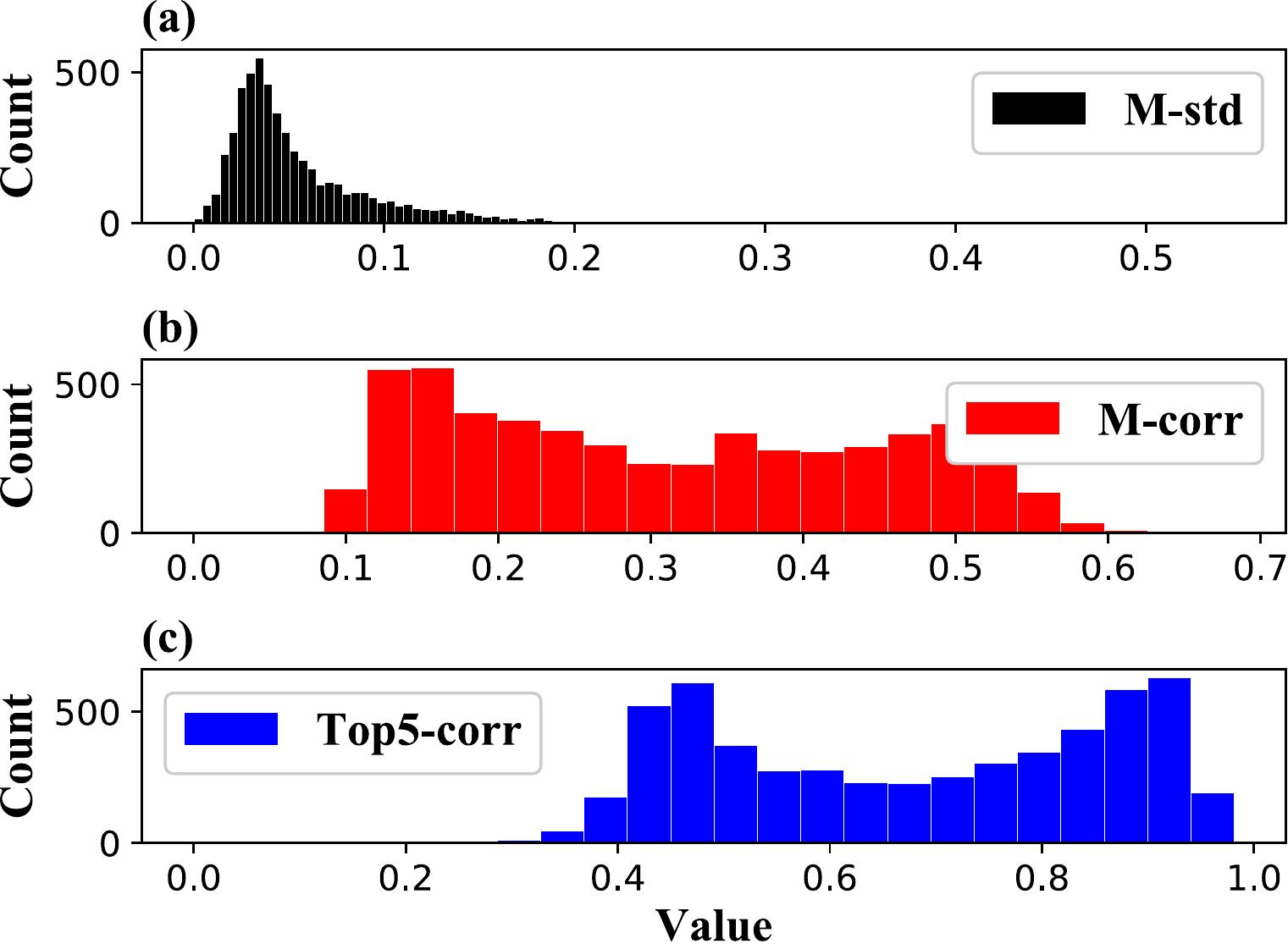}
\caption{Distributions of M-std, M-corr and Top5-corr of features in VGGNet trained on CIFAR 10 dataset.}
\label{fig:hist}
\end{figure}

\begin{figure}[tb]
\centering
\includegraphics[width=0.65\linewidth]{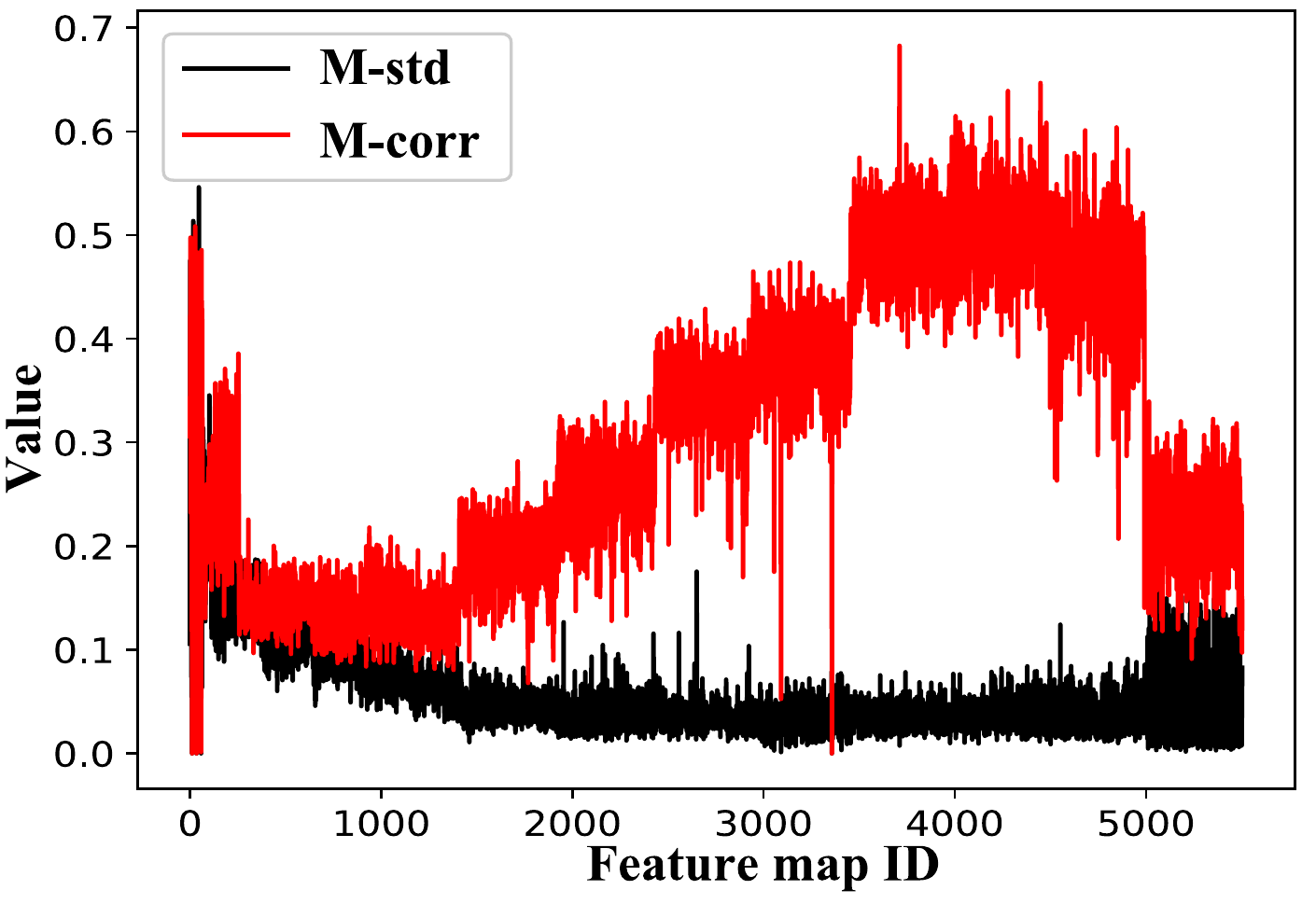}
\caption{M-std and M-corr of all feature maps in VGGNet trained on CIFAR 10 dataset.}

\label{fig:std_corr}
\end{figure}
To measure the diversity and similarity of feature maps, we adopt two metrics:

\begin{itemize}[leftmargin=*]
  \item Mean standard deviation (M-std).
  The mean standard deviation of each feature map is defined as:
  \begin{equation} \label{equ:std}
      \text{M-std}(\mathbf{X}_{i,j}) = \frac{1}{T} \sum_{m=1}^T 
      \sqrt{\frac{\sum_{p=1}^{W_i H_i } (x_p^{m}- \overline{x}^m)^2}{W_i H_i -1}} \ ,
  \end{equation}

where $x_p^m$ is the $p_{th}$ element in $\mathbf{x}_{i,j}^m \in \mathbb{R}^{1\times W_i H_i}$ of a flat feature map $\mathbf{X}_{i,j}^m \in \mathbb{R}^{W_i\times H_i}$, $m$ ($m \leqslant T$) represents the sample index, and $\overline{x}^m$ is the mean of $\{x_p^m\}$. Smaller M-std value means the corresponding feature map has less heterogeneity. This lower information diversity contributes more inadequate to further feature extraction.

  \item Mean cosine similarity (M-corr). We use the cosine similarity to measure the relevance between feature maps. Since the dimensions of feature maps vary in different layers, we compute the feature similarity within the same layer. The M-corr can be computed as:
  \begin{equation}
      \text{M-corr}(\mathbf{X}_{i,j}) = \frac{1}{T} \sum_{m=1}^T \frac{\sum_{p=1}^{N_i}  \left | { \text{cos}(\mathbf{x}_{i,j}^m,\mathbf{x}_{i,p}^m)}\right |}{N_i}  \ . 
  \end{equation}{}
\end{itemize}
The feature map with a larger M-corr value tends to have high similarity with each feature in its layer.
Comparing with M-corr, mean Top-k cosine similarity (Topk-corr) is an another commonly used metric that can also reflect the similarity characteristic of feature maps. The Topk-corr of a feature map $\mathbf{X}_{i,j}$ is $ \frac{1}{T} \sum_{m=1}^T \frac{\sum_{p=1}^{N_i} A_{i,p} \cdot \ \left | {\text{cos}(\mathbf{x}_{i,j}^m,\mathbf{x}_{i,p}^m)}\right |} {k}$. $A_{i,p}=1$ if $\mathbf{x}_{i,p}^m$ is among the Top-k cosine values of $\mathbf{x}_{i,j}^m$, $A_{i,p}=0$ otherwise.

To highlight the characteristics of feature maps, we train a VGGNet model on the CIFAR10 dataset and obtain the statistical information of feature maps.
Figure~\ref{fig:hist} gives an illustration of the overall distributions of M-std, M-corr, and Topk-corr. As shown in Figure~\ref{fig:hist}a, there are nearly a half number of feature maps with M-std less than 0.05, which reveals that these feature maps may not contain much information. Figure~\ref{fig:hist}b indicates there are around half of the feature maps that have M-corr value over 0.3. 
Figure~\ref{fig:hist}c shows that there are about a quarter of feature maps with Top-5 cosine similarity values exceed 0.8, even 0.9. Similar feature maps can be treated as redundant features, making less contribution to the network. These metrics demonstrate there exist redundant feature maps in CNNs.

The details of the statistic value of each feature map are shown in Figure~\ref{fig:std_corr}. We can see that most of the M-std values become lower with the increase of layer depth and most of the M-corr values enhance as the number of channels gets larger. M-std and M-corr have a weak negative correlation between each other in this case. Their Pearson correlation~\cite{galton1886regression} is -0.38. These two criteria can complement each other when selecting important features.

\subsection{Filter Pruning}
We aim to prune redundant filters of deep CNNs in a simple but effective scheme. The central idea of our method has two stpdf of feature map selections (see Figure~\ref{fig:pip}): diversity-aware feature selection (DFS) and similarity-aware feature selection (SFS).
After obtaining a pre-trained CNN model, we first compute the M-std values for all feature maps. Then we prune feature maps with the smallest values and save the unpruned feature maps. Finally, we calculate the cosine similarity among the unpruned feature maps and prune those features with high similarity for further compression. The overall filter pruning scheme is illustrated in Algorithm \ref{alg:all}.

\begin{algorithm}[tb]
\caption{Our proposed filter pruning scheme}
\label{alg:all}
\textbf{Input}: Sample data $\{\hat{x}_i\}_{i=1}^T$, model $\mathbf{M}$, threshold $\nu$ \\
\textbf{Output}: Selected filter subset $\hat{\mathcal{F}}$
\begin{algorithmic}[1] 
\STATE Construct feature maps $\{\mathbf{X}_i^m\}_{i=1}^L$ for each data sample
\STATE Compute M-std $\{std\}$ from Equation \ref{equ:std} for each feature using $\{\mathbf{X}_i^m\}_{i=1}^L$ 
\STATE Let $\hat{\mathbf{X}} \leftarrow  \varnothing$, $\beta \leftarrow mean (\{std\})$
\FOR{$i \in \{1,2,...,L\}$}
\STATE Find $\hat{\mathbf{X}}_i$ according to Equation \ref{equ:select_std}
\STATE Obtain $\widetilde{\mathbf{X}}_i$ using Algorithm \ref{alg:second-step}, $\widetilde{\mathbf{X}}_i \leftarrow $ SFS($\hat{\mathbf{X}}_i, \nu$)
\STATE Let $\hat{\mathbf{X}} \leftarrow \hat{\mathbf{X}} \cup \widetilde{\mathbf{X}}_i$
\ENDFOR
\STATE Find $\hat{\mathcal{F}}$ according to $\hat{\mathbf{X}}$
\end{algorithmic}
\end{algorithm}

We adopt M-std as the diversity criteria. Specifically, in the $i$-th layer, the selected feature $\hat{\mathbf{X}}_i$ at DFS is:
\begin{equation}\label{equ:select_std}
   \hat{\mathbf{X}}_i = \{\mathbf{X}_{i,j} \ | \ \text{M-std}( \mathbf{X}_{i,j}) \geqslant \beta \}, \ j =1,2,..., N_i \ ,
\end{equation}
where $\beta$ is the hyper-parameter which is chosen according to percentiles of M-std values of all feature maps, making our method a global pruning across all layers. 

To select a subset of features with lower correlations, we employ a direct way to delete redundant features with high similarity. In particular, we compute the cosine similarity among all the features of $\hat{\mathbf{X}}_i$, which can form a correlation set $\mathbf{s}_i = \{s_{i_{j,p}}\}$ ,
\begin{equation}\label{equ:cosine}
     s_{i_{j,p}}=\frac{1}{T} \sum_{m=1}^T \frac{ \ | \text{cos}(\mathbf{x}_{i,j}^m,\mathbf{x}_{i,p}^m)\ | }{T}, \mathbf{x}_{i,\ast} \in \hat{\mathbf{X}}_i. 
\end{equation}
We find the largest value in $\mathbf{s}_i$ and its corresponding feature pair, then we save one of the pair as a reference feature $\mathbf{x}_{i,r}$\,. As a result, we can safely delete features whose similarity with $\mathbf{x}_{i,r}$ is bigger than a pre-defined threshold $\nu$ because these features could be replaced by $\mathbf{x}_{i,r}$. SFS is summarized in Algorithm \ref{alg:second-step}.

\begin{algorithm}[tb]
\caption{Similarity-aware feature map selection (SFS)}
\label{alg:second-step}
\textbf{Input}: Feature map set ${\mathbf{X}}_i$, threshold $\nu$ \\
\textbf{Output}: Selected feature subset $\mathcal{B}_i$
\begin{algorithmic}[1] 
\STATE Initialize $\mathcal{B}_i \leftarrow  \varnothing $
\STATE Form the correlation set $\mathbf{s}_i$ according to Equation \ref{equ:cosine}
\STATE Find the max value $max$ in $\mathbf{s}_i$ 
\WHILE{$max > \nu$}
\STATE Find $\mathbf{X}_{i,r}$ and $\mathbf{X}_{i,c}$ have the $max$ similarity value
\STATE Let $\mathcal{B}_i \leftarrow \mathcal{B}_i \cup \mathbf{X}_{i,r} $
\FOR{$\mathbf{X}_{i,j} \in \mathbf{X}_{i}$}
\IF{$s_{i_{r,j}} > \nu $}
\STATE Remove $s_{i_{r,j}}$ from $\mathbf{s}_i$ and remove $\mathbf{X}_{i,j}$ from $\mathbf{X}_{i}$
\ENDIF
\ENDFOR
\STATE Find $max$ in $\mathbf{s}_i$ 
\ENDWHILE
\STATE Let $\mathcal{B}_i \leftarrow \mathcal{B}_i \cup \mathbf{X}_{i} $
\STATE \textbf{return} $\mathcal{B}_i$
\end{algorithmic}
\end{algorithm}

\subsection{Pruning for Multiple Branch Networks}

\begin{figure*}[tb]
\centering
  \begin{subfigure}[b]{0.2\textwidth}
  \includegraphics[height=3cm]{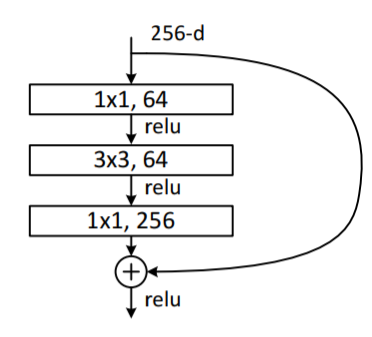}
     \caption{bottleneck} \label{fig:bottleneck}
  \end{subfigure}
  \begin{subfigure}[b]{0.25\textwidth}
    \includegraphics[width=\textwidth]{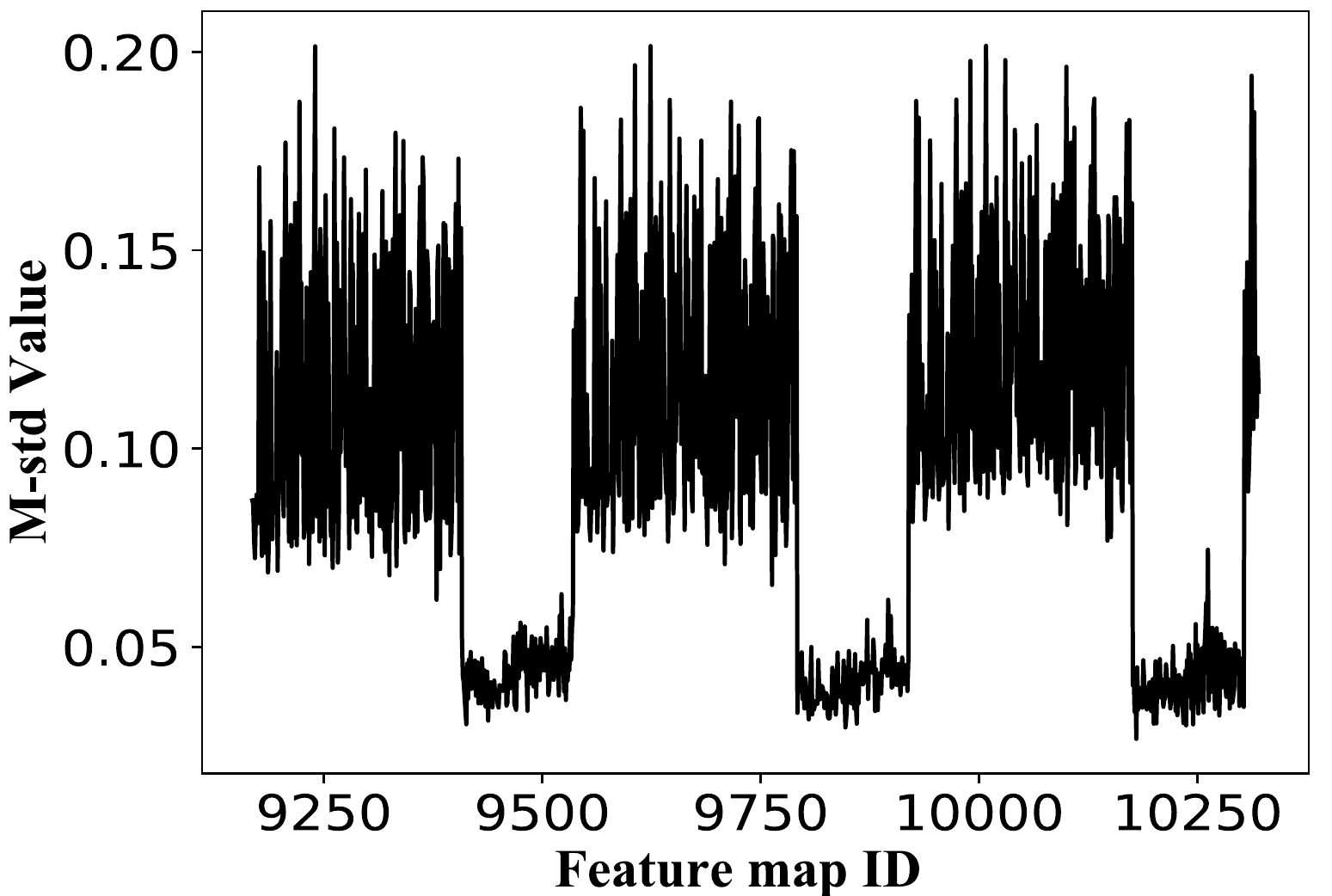}
     \caption{M-std of part feature maps.} \label{fig:resnet_sta}
  \end{subfigure}
  \begin{subfigure}[b]{0.25\textwidth}
    \includegraphics[width=\textwidth]{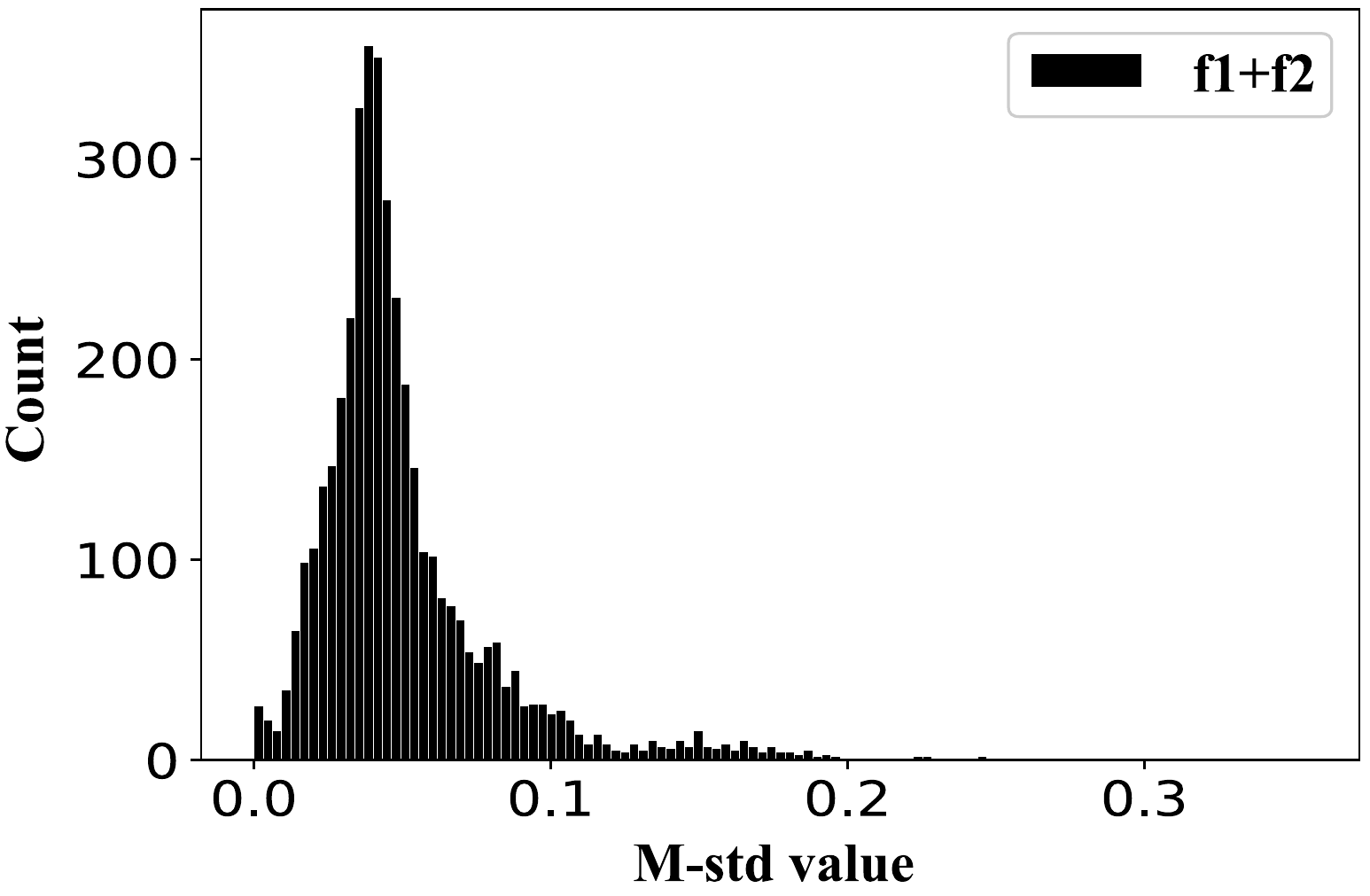}
    \caption{M-std distribution of first two layers within sequential branches.} \label{fig:resnet_12_std}
  \end{subfigure}
  \begin{subfigure}[b]{0.25\textwidth}
    \includegraphics[width=\textwidth]{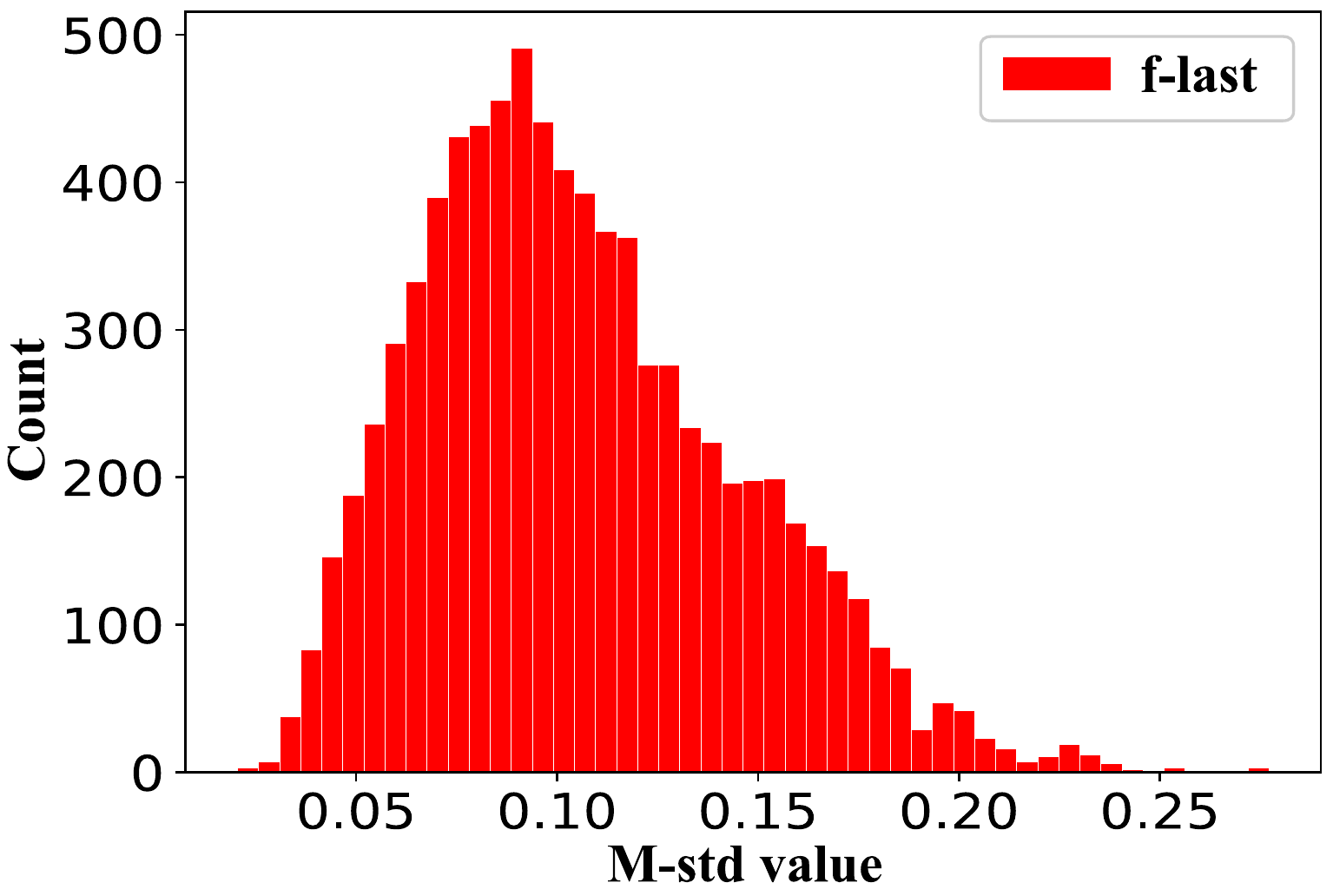}
    \caption{M-std distribution of combination layers.} \label{fig:resnet_3_std}
  \end{subfigure}
  \caption{\textit{Bottleneck} and M-std values of ResNet-164 on CIFAR-10.} 
  \label{fig:resnet_all}
\end{figure*}

The multiple branch networks illustrate a kind of CNNs that the output of one layer may be the input of multiple subsequent layers, which are more complicated to prune than single branch networks. For instance, ResNet~\cite{resnet} is a representative example of multiple branch networks, which has a sequential branch and a shortcut branch. The outputs of these two branches will conduct an element-wise addition operation. Since the outputs require equal channel dimensions, this makes pruning ResNet more difficult.

We use two separate feature map selection processes for the sequential branch and the shortcut branch, respectively. 
The features except for the last layer within all sequential branches compose one group, the results after branches combination form another group. Filter pruning is operated on each group individually. 
These two separate filter pruning strategies are inspired by the statistic information of feature maps in PreResNet~\cite{he2016identity}. Figure~\ref{fig:bottleneck} gives an example of the bottleneck architecture, which is one of the multiple-branch building blocks of ResNet. This bottleneck includes three layers with $1\times1$, $3\times3$, and $1\times1$ convolutional filters. The element-wise addition is performed channel by channel on two output feature maps of sequential and shortcut branches. We train a PreResNet-164 on the CIFAR10 dataset and compare the statistical information of feature maps between those in the first two layers of sequential branches (f1+f2) and those after the additional operation (f-last). Figure~\ref{fig:resnet_sta} shows the M-std values of part of the feature maps. The overall pattern of M-std for all feature maps is similar to Figure~\ref{fig:resnet_sta}. We can clearly see that all M-std values form two groups, the upper part corresponds to f1+f2, and the lower one represents f-last. Besides, the distribution of M-std of f1+f2 is mainly from 0 to 0.1, which is shown in Figure~\ref{fig:resnet_12_std}. From Figure~\ref{fig:resnet_3_std}, we could conclude that the M-std of f-last is about 0.05 to 0.2. The percentile of f1+f2 is higher while the one of f-last is lower, which motivates us to use two different pruning thresholds. Thus, we utilize two feature map selection processes for f1+f2 and f-last layer, respectively.

\section{Experiments}
\begin{table}[tb]
\small
\centering
\begin{tabular}{lllll}  
\toprule
Dataset  & Method  & Acc.(\%) & FLOPs $\downarrow$(\%) & Para.$\downarrow$(\%) \\
\midrule
C10     & VGGNet  & 93.74  & 0.0  & 0.0 \\
        & {L1-Prune}& 93.12  & -  &88.5 \\
         &{N-Slim}$\ast$ & 93.80  &51.1  & 88.5 \\
         &{PFGM}$\ast$  & 94.0  &35.9    & - \\
         &\textbf{\NAME}  &\textbf{94.05}   & \textbf{56.3} & \textbf{90.7} \\
\hline
C100 & VGGNet  & 73.41  & 0.0  & 0.0  \\
     & {L1-Prune} & 71.64  & -  &76.0 \\
     &{N-Slim}$\ast$  & 73.48  & 37.1 &75.1 \\
     &\textbf{\NAME}    &\textbf{73.69}  &\textbf{45.0} &\textbf{75.9} \\
\bottomrule
\end{tabular}
\caption{Results of pruned VGGNet on CIFAR dataset. 
C10 and C100 mean the CIFAR~10 and CIFAR~100, respectively.
Acc. is the classification accuracy, and Para. is short for parameters. The $\downarrow$ is the drop percent between the pruned model and the original model, the smaller, the better. Results with $\ast$ are got from original papers. $-$ denotes the results are not reported.}
\label{tab:VGG}
\end{table}

\subsection{Experimental Setting}
\paragraph{Dataset.} We perform experiments on publicly available datasets. CIFAR10 and CIFAR100~\cite{krizhevsky2009learning} are two widely used datasets with $32\times32$ colour natural images. They both contain $50,000$ training images and $10,000$ test images with 10 and 100 classes respectively. The data is normalized using channel means and standard deviations. And the data augmentation approach we used is consistent with~\cite{liu2017learning}. ILSVRC-2012 is a large-scale dataset with 1.2 million training images and $50,000$ validation images of 1000 classes. Following the common training procedure in~\cite{liu2017learning,he2019filter}, we adopt the same data augmentation approach and report the single-center-crop validation error of the final model.

\paragraph{Network models.} We test the performance of our pruning method on several famous CNN models. VGGNet is a remarkable single branch network which is widely used for computer vision task. ResNet~\cite{resnet} and PreResNet~\cite{he2016identity} are two popular multiple branch network. MobileNet~\cite{howard2017mobilenets} is a compact network designing for effective use on mobile devices.

\paragraph{Configuration.} We train or fine-tune all the networks using SGD. For CIFAR, we set the mini-batch size to 64, epochs to 160 with a weight decay of 0.0015 and Nesterov momentum~\cite{sutskever2013importance} of 0.9. For ILSVRC-2012, we use the pre-trained ResNet-50 released by Pytorch. We train MobileNet for 60 epochs with a weight decay of 0.0015. The pruning ratio is determined by two factors, one is a percentile among M-std and the other is the threshold for SFS, i.e. 40\% for DFS, 0.85 for SFS. 

\subsection{Compared Algorithms}
\begin{itemize}[leftmargin=*]
    \item \textbf{L1-Prune}\footnote{The result of L1-Prune is obtained from~\cite{liu2017learning}.}~\cite{li2016pruning} uses the $l1$-norm of filters as the important measurement.
    \item \textbf{ThiNet}~\cite{luo2017thinet} is a feature-map based method that selects the filter subset reconstructing the next layer.
    \item \textbf{N-Slim}~\cite{liu2017learning} gets the importance of each filter during the training process according to the batch-normalization scaling factors.
    \item \textbf{PFGM}~\cite{he2019filter} prunes redundant filters utilizing geometric correlation among filters in the same layer.
\end{itemize}

\subsection{Experimental Results}

\begin{table}[tb]
\small
\centering
\begin{tabular}{lllll}  
\toprule
Dataset  & Method  & Acc.(\%) & FLOPs $\downarrow$(\%) & Para.$\downarrow$(\%) \\
\midrule
C10  & PreResNet  & 94.86  &0.0  &0.0 \\
         &{N-Slim}$\ast$ & 94.73  & 44.9  & 35.2  \\
         &\textbf{\NAME}     & \textbf{94.93}  & \textbf{56.1} & \textbf{40.5} \\
\hline
C100  & PreResNet  & 76.88  & 0.0   & 0.0 \\
          &{N-Slim}$\ast$ & 76.09  & 50.6  & 29.7 \\
          &\textbf{\NAME}   &\textbf{76.18}   & \textbf{53.4} & \textbf{35.9} \\
\bottomrule
\end{tabular}
\caption{Comparisons of pruning PreResNet on CIFAR dataset.}
\label{tab:Pre-resnet}
\end{table}

\begin{table}[tb]
\centering
\small
\begin{tabular}{llll}
\hline
Method  & Acc.(\%) & FLOPs $\downarrow$(\%) & Para.$\downarrow$(\%) \\
\hline
ResNet      & 76.15   & 0.0  & 0.0 \\
ThiNet      & 72.04   & 40.5  & 33.7 \\
PFGM $\ast$ & 75.03   & 42.2    & 39.6 \\
\textbf{\NAME}  &71.05   &40.4 &47.8\\
\hline
\end{tabular}
\caption{Comparisons of pruning ResNet on ILSVRC-2012.}
\label{tab:Resnet}
\end{table}

\paragraph{Single branch network.} We first prune the trained VGGNet on CIFAR10 and CIFAR100. We compare the performance of our method with the state-of-the-art methods. Table~\ref{tab:VGG} lists the results of the classification accuracy, the reduction of parameters, and the decrease in FLOPs, respectively. Although all approaches can reduce the model size with limited accuracy drop, our method has the highest compression ratio. The scalar factors used in N-Slim are not powerful enough for compression since it does not consider relationships among features. Although PFGM can achieve satisfactory accuracy, it has low pruning ratio, since it neglects the feature diversity.
On the other hand, our method considers both the diversity and similarity of features maps, which benefits the performance improvements of our method over other methods. When pruning 90.7\% of parameters and 56.3\% of FLOPs of VGGNet trained on CIFAR~10, our method surprisingly increases the accuracy by 0.31\%. One possible reason is that our method reduces the unnecessary parameters that cause the overfitting of the original model.

\paragraph{Multiple branch network.} We prune PreResNet-164 on CIFAR and ResNet-50 on ILSVRC-2012, respectively. The results are reported in Table~\ref{tab:Pre-resnet} and Table~\ref{tab:Resnet}. From the results, we can observe that even with multiple branch networks, our method can still compress the model to a satisfactory extent. After SFS and DFS, our method reduces up to 40\% of parameters and 56.1\% of FLOPs for PreResNet on CIFAR~10 while maintaining the accuracy as high as 94.93\%. 

\paragraph{Compact designed network.} To further illustrate the generalization of our method, we prune MobileNet on both CIFAR and ILSVRC-2012 datasets. The performance of different prune ratios is given in Table~\ref{tab:mobilenet}. With the increase of the compression ratio, the accuracy of the pruned model drops gradually. Although the pruned model reduces 91.6\% of parameters and 83.7\% FLOPs, its accuracy as high as 93.17\% on CIFAR~10.

\begin{table}[tb]
\centering
\small
\begin{tabular}{lllll}  
\toprule
Dataset  & Method  & Acc.(\%) & FLOPs $\downarrow$(\%) & Para.$\downarrow$(\%) \\
\midrule
C10  & MobileNet   &93.71   & 0.0 & 0.0 \\
      &{R1} &93.91   & 47.1 & 66.7 \\
      &{R2} &93.86   & 65.8 & 80.7 \\
      &{R3} &93.17   & 83.7 & 91.6 \\
\hline
C100 & MobileNet   &74.19   & 0.0    & 0.0   \\
      &{R1} &75.40   & 29.3 & 43.8 \\
      &{R2} &74.77   & 47.8 & 58.3 \\
      &{R3} &72.73   & 62.6 & 68.3\\
\hline
ILSVRC   & MobileNet  &68.43 & 0.0 & 0.0  \\
         &{R1} & 67.54   &37.06 & 40.18 \\
         &{R2} & 61.20  &61.49  & 67.13 \\
\bottomrule
\end{tabular}
\caption{Performance of pruned MobileNet on CIFAR and ILSVRC-2012. R$k$ denotes different compression step with $\beta$ is 25\% quantile of M-std and $\nu$=0.85. R1 is the pruning based on the pre-trained model. R2 prunes the result of R1. R3 is the final pruning based on result of R2.
}
\label{tab:mobilenet}
\end{table}

\begin{figure}[tb]
\centering
\includegraphics[width=0.85\linewidth]{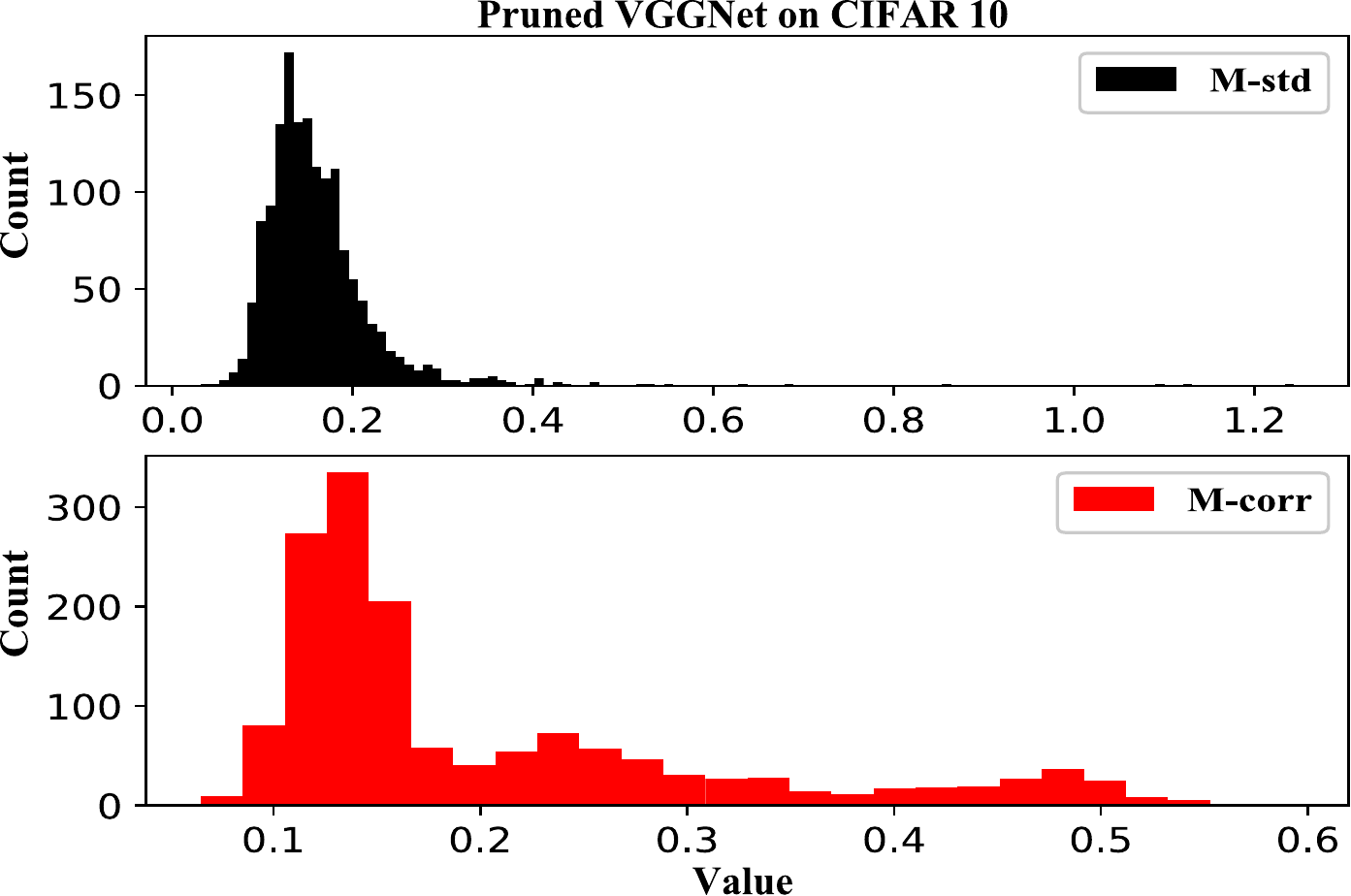}
\caption{Distributions of M-std and M-corr of all feature maps in pruned VGGNet trained on CIFAR 10 dataset.}
\label{fig:after_hist}
\end{figure}

\begin{figure}[tb]
\centering
\includegraphics[width=1\linewidth]{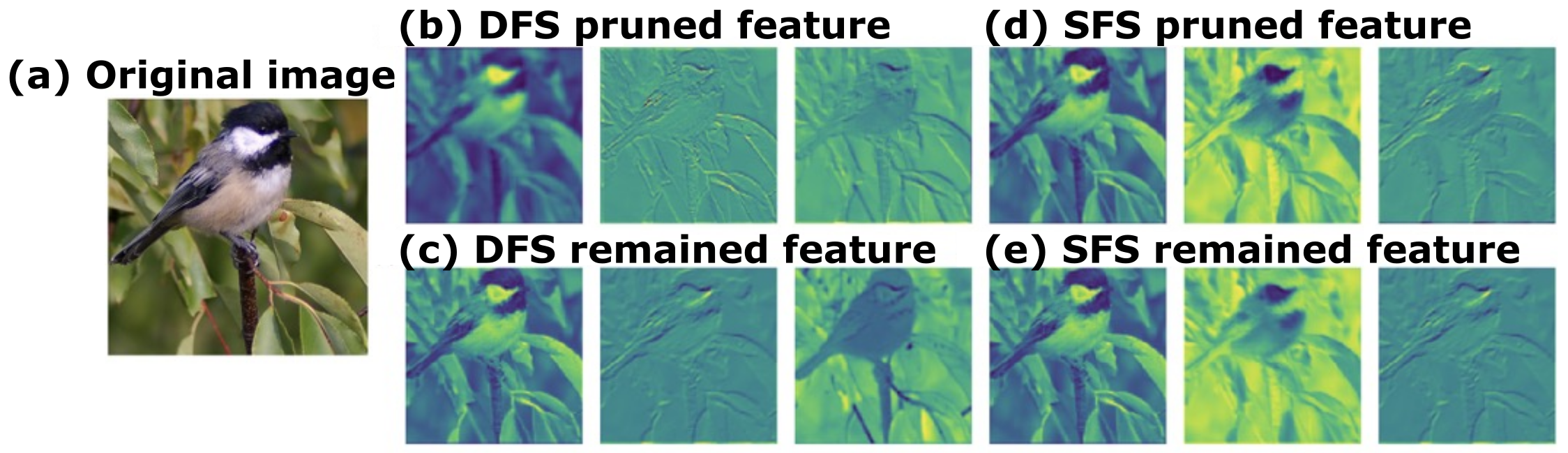}
\caption{A demonstration of pruned and remained feature maps.}
\label{fig:show_feature}
\end{figure}

\paragraph{The efficiency of feature map selection.} The purpose of our two feature selections is to extract more diverse and less similar feature maps. We prune a VGGNet trained on CIFAR~10 and plot the statistic information of feature maps in Figure~\ref{fig:after_hist}.
It can be observed that most M-std values of feature maps are concentrated between 0.1 and 0.2, which is higher than the original model (i.e. 0.05 in Figure~ \ref{fig:hist}a). In addition, the M-corr values are almost smaller than 0.3, which is much lower compared with the original model (i.e. Figure~ \ref{fig:hist}b). These indicate the remained features have higher diversities and fewer similarities. Although there still exist feature maps with M-std values smaller than 0.05 and M-corr values bigger than 0.5, their percentage is quite small. As a result, these feature maps can keep the generalization ability of the model.

\paragraph{Visualization.} We further visualize the pruned and remained feature maps to show the effectiveness of our approach. Figure~\ref{fig:show_feature} shows part of the feature maps of the first convolutional layer in ResNet-50 trained on LSVRC-2012. Each feature map in Figure~\ref{fig:show_feature}b expresses limited information variance (e.g. ambiguous or no texture) compared with Figure~\ref{fig:show_feature}c.
The feature maps in Figure~\ref{fig:show_feature}d are nearly the same as Figure~\ref{fig:show_feature}e. Therefore, SFS prunes one of them to keep fewer similar features.

\section{Conclusion}
In this study, we investigate the statistical information of feature maps in CNN for analyzing the diversity and similarity. We propose two feature map selections, namely DFS and SFS, for removing redundant filters. The pruning method we proposed can significantly compress the size of CNNs and decrease the computational cost with almost no accuracy loss. To further validate the effectiveness of our proposed method, different pruning ratio strategies can be evaluated in the future. We will also explore more tasks, such as object detection and text classification.


\bibliographystyle{named}
\bibliography{ijcai20}

\end{document}